\documentclass[11pt]{article}

\usepackage[preprint]{acl}  % "final": camera-ready, "preprint": non-anonymous with page numbers

\usepackage{times}
\usepackage{latexsym}
\usepackage[T1]{fontenc}
\usepackage[utf8]{inputenc}
\usepackage{microtype}
\usepackage{inconsolata}
\usepackage{graphicx}
\usepackage{amsmath}
\usepackage{booktabs}
\usepackage{array}
\newcolumntype{L}[1]{>{\raggedright\arraybackslash}p{#1}}
\usepackage{hyperref}
\usepackage{color}

\title{Controlling the Diversity--Quality Trade-Off of LLMs by Merging Models Before and After Alignment}

\author{
	Yasuaki Sumita \\
	Graduate School of Informatics, \\
	Kyoto University, Kyoto, Japan \\
	\texttt{sumita.yasuaki.73a@st.kyoto-u.ac.jp} \\\And
	Tomoharu Iwata \\
	NTT Communication Science Laboratories, \\
	NTT, Inc., Kyoto, Japan \\
	\texttt{tomoharu.iwata@ntt.com} \\\AND
	Toshiyuki Tanaka \\
	Graduate School of Informatics, \\
	Kyoto University, Kyoto, Japan \\
	\texttt{tt@i.kyoto-u.ac.jp}
}

\begin{document}

\maketitle

\begin{abstract}
	% 200 words.
	Alignment is known to change the output characteristics of large language models (LLMs).
	Base models tend to generate more diverse outputs, whereas aligned models tend to produce higher-quality outputs.
	The appropriate balance between the two depends on the task.
	Existing methods either require costly fine-tuning or, like temperature sampling, mainly improve lexical diversity rather than semantic diversity.
	We introduce a depth-aware FrankenMerging approach for controlling the diversity--quality trade-off between an aligned model and its base model without additional training.
	This approach combines the lower layers of an aligned model with the higher layers of its base model based on the hypothesis that lexical diversity is more closely associated with lower layers and semantic diversity with higher layers.
	Experiments showed that varying the number of layers taken from each model adjusts both diversity and quality.
	In most cases, the merged models exhibited higher semantic diversity than the aligned model, and at some split positions, they showed higher lexical diversity than both original models.
	They also demonstrated a better diversity--quality trade-off than temperature sampling in most settings.
	A supporting ablation, in which we reversed the merging order, yielded results largely consistent with the hypothesis.
\end{abstract}

\section{Introduction}\label{sec:introduction}
Large language models (LLMs) are used in a wide range of text generation tasks.
LLMs are divided into two types: base models that have undergone only pretraining, and aligned models that have been fine-tuned to conform to human preferences.
These two types of models have different output characteristics: base models tend to generate more diverse outputs, whereas aligned models tend to produce outputs of higher quality~\cite{kirk2024understanding,shypula2025evaluating}.
The relative importance of these two characteristics depends on the task and the purpose of use.
For example, while output quality is primarily emphasized in summarization, output diversity is essential in story generation.
\citet{xu2025echoes} demonstrate that LLMs tend to generate only plots with similar developments in story generation.
Therefore, it is necessary to balance the diversity and quality of LLM outputs according to the task.

Existing methods for enhancing output diversity have two limitations.
First, some methods~\cite{wang2025diversityoriented,lanchantin2025diverse,chung2025modifying} fine-tune LLMs to improve diversity, but these have high computational costs as they require additional training.
Second, there are methods that do not require training, such as temperature sampling~\cite{ackley1985learning}, but these primarily enhance lexical diversity, which is surface-level variation in vocabulary, and contribute little to semantic diversity, which is variation in content~\cite{tevet2021evaluating,guo2025benchmarking}.

To address these limitations, this study aims to control the diversity--quality trade-off between an aligned model and its base model without additional training.
We evaluate diversity in terms of both lexical diversity and semantic diversity.
To this end, we introduce a depth-aware FrankenMerging approach.
FrankenMerging~\cite{goddard2024arcees} is a model merging methodology that constructs new models by combining partial layers from multiple input models.
Our approach selects the source model of each layer according to its depth and combines the lower layers of an aligned model with the higher layers of its base model.\footnote{In this paper, lower layers refer to the layers closer to the input, while higher layers refer to the layers closer to the output.}
This approach is motivated by existing findings that the roles of LLM layers differ with their depth~\cite{liu2024fantastic}.
On the basis of these findings, we hypothesize that lexical diversity is more closely associated with lower layers, whereas semantic diversity is more closely associated with higher layers.
Under this hypothesis, it is expected that by utilizing the higher layers extracted from the base model, its semantic diversity can be brought into the merged model, and that the balance between diversity and quality can be adjusted by varying the split position.

To evaluate our approach, we conducted experiments across four tasks: translation, summarization, story generation, and dialogue generation.
In these experiments, we evaluated the output diversity and quality of the merged models for different numbers of layers taken from each model.
Furthermore, we compared the diversity--quality trade-off of our approach with that of temperature sampling.
Additionally, as a supporting ablation for the hypothesis underlying our approach, we also assessed how reversing the model merging order affects output diversity.

The contributions of this study are as follows:
\begin{itemize}
	\item We introduce a depth-aware FrankenMerging approach for controlling the diversity--quality trade-off between an aligned LLM and its base model and experimentally demonstrate that varying the number of layers taken from each model adjusts both diversity and quality, and that the approach provides a better diversity--quality trade-off than temperature sampling in most settings.
	\item We empirically investigate how layer depth relates to lexical and semantic diversity.
	      Our results, together with those of a supporting ablation in which we reverse the merging order, are largely consistent with the hypothesis above.
\end{itemize}

\section{Related Work}\label{sec:related-work}
\subsection{Diversity in Natural Language Generation}
Some studies evaluate linguistic diversity in natural language generation.
\citet{tevet2021evaluating} reviewed existing diversity assessment metrics and classified them as follows: $n$-gram-based metrics like Distinct-$n$~\cite{li2016diversitypromoting} are primarily used for evaluating lexical diversity, while embedding-based metrics such as the distance between sentence embeddings computed by SentenceBERT~\cite{reimers2019sentencebert} are better suited for assessing semantic diversity.
\citet{zhang2021trading} identified the trade-off between diversity and quality in natural language generation.
They demonstrated that texts generated with high likelihood do not necessarily correspond to high-quality outputs according to human evaluation.

Several studies investigate methods for evaluating linguistic diversity in text generated by LLMs.
\citet{guo2025benchmarking} developed a framework for assessing the linguistic diversity of LLM outputs across lexical, syntactic, and semantic dimensions.
They measured diversity across outputs for different inputs and reported that instruction-tuned models show higher lexical diversity but lower semantic diversity than their base models.
\citet{kirk2024understanding} examined how fine-tuning and alignment affect the generalization performance and output diversity of LLMs.
They found that alignment improves generalization but yields lower output diversity than supervised fine-tuning, and that this reduction is smaller for diversity across inputs than for diversity among outputs for a single input.
\citet{shypula2025evaluating} also showed that preference-tuned models produce outputs with lower lexical and semantic diversity for the same input than supervised fine-tuned or base models.
These findings suggest that the effect of alignment on output diversity depends on the aspect of diversity and on the evaluation protocol.
Nevertheless, the reduction in some aspects of diversity can pose challenges for tasks requiring diverse outputs, such as story generation and dialogue generation.
In this study, we aim to control the diversity--quality trade-off between an aligned LLM and its base model.

\subsection{Improving Output Diversity in LLMs}
Improving output diversity is important across various tasks such as story generation~\cite{xu2025echoes}, data augmentation~\cite{guo2024curious}, and dialogue generation~\cite{stasaski2022semantic}.
For example, in data augmentation, low-diversity LLM outputs may lead to reduced generalization performance because the augmented data do not sufficiently reflect the diversity of the original dataset.

Research aimed at enhancing the diversity of LLM outputs primarily employs two approaches.
The first approach develops decoding algorithms, which determine the next token on the basis of the given probability distribution.
The most representative method is temperature sampling~\cite{ackley1985learning}, which flattens the token probability distribution to increase lexical diversity.
Other techniques include top-$k$ sampling~\cite{fan2018hierarchical}, locally typical sampling~\cite{meister2023locally}, and min-p sampling~\cite{minh2025turning}.
These sampling methods, however, have been shown to primarily improve word-level diversity and to be ineffective for producing semantically diverse outputs~\cite{tevet2021evaluating,guo2025benchmarking}.

The second approach involves improving diversity through fine-tuning of LLMs.
For example, \citet{wang2025diversityoriented} proposed a method that enhances output diversity by incorporating diverse paraphrased examples into the dataset used for direct preference optimization (DPO)~\cite{rafailov2023direct}.
\citet{lanchantin2025diverse} presented a method that constructs a preference dataset, wherein high-diversity outputs are incentivized with positive feedback, while low-diversity outputs are penalized during DPO training.
\citet{chung2025modifying} further proposed a modified DPO variant that weights the loss by the cosine distance between the model's responses and reference answers.
Nevertheless, all these approaches involve modifying LLM fine-tuning methods, which results in high computational costs.
To address this issue, this study introduces an approach that controls the diversity--quality trade-off between an aligned model and its base model through model merging, which requires no additional training.

\subsection{Model Merging}
Model merging is a methodology that integrates parameters from multiple pre-trained models into a single model with the aim of achieving higher performance~\cite{goddard2024arcees}.
One significant merit of model merging is the ability to construct high-performance models without requiring computationally intensive processes like fine-tuning.

Various model merging methods have been proposed.
Model soups~\cite{wortsman2022model} averages parameters of models trained with different hyperparameters and achieves higher performance than any individual base model.
Git Re-Basin~\cite{ainsworth2023git} extends this to models with different initializations by solving a neuron matching problem.
Evolutionary model merging~\cite{akiba2025evolutionary} uses evolutionary computation to optimize which layers to merge and in what order.
\citet{goddard2024arcees} proposed FrankenMerging via the MergeKit framework, which constructs new models by selectively combining layers from multiple source models without modifying their parameters.

Nevertheless, these studies have focused only on improving the performance of merged models.
No previous work has addressed model merging methods specifically aimed at enhancing output diversity.
Furthermore, although some of these methods select or combine layers, they do so to improve task performance and do not consider how layers at different depths relate to output diversity.
In this study, we introduce a depth-aware FrankenMerging approach that takes into account how layers at different depths relate to output diversity.

\section{Approach}\label{sec:approach}
This study introduces a depth-aware FrankenMerging approach that merges models before and after alignment into a new model.
The approach selects the source model of each layer according to its depth.
By combining these models, we aim to control the balance between the diversity of the base model and the quality of the aligned model.

\begin{figure}[tbp]
	\centering
	\includegraphics[width=\columnwidth]{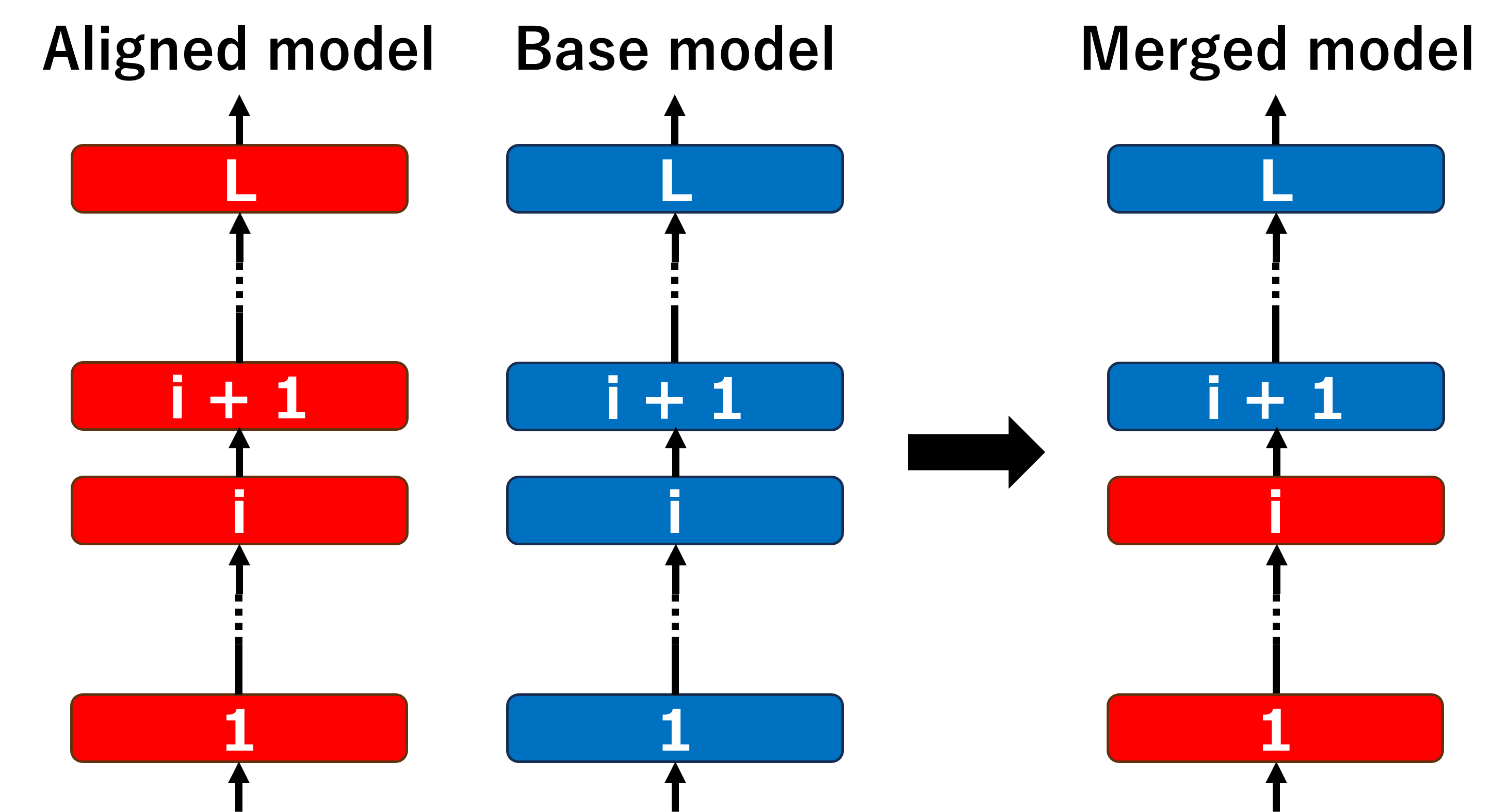}
	\caption{Overview of our approach: We sequentially combine layers from the 1st to the $i$-th layer of an aligned model A with layers from the $(i+1)$-th to the $L$-th layer of its base model B, then output the resulting model as a new one.}
	\label{fig:proposed-method}
\end{figure}

The procedure of our approach is as follows.
First, we employ two inputs: model A, which has been fine-tuned with alignment, and base model B, which is the model before fine-tuning with alignment.
Since both A and B share the same architecture with the same number of layers, we let $L$ denote this number and specify a hyperparameter $i\in\left\{1,2,\dots,L-1\right\}$.
We then sequentially combine the layers from the first to the $i$-th layer of the aligned model A with the layers from the $\left(i+1\right)$-th to the $L$-th layer of the base model B, and output the resulting model as the new one.
The overview of our approach is shown in Figure \ref{fig:proposed-method}.

Our approach draws inspiration from studies on how the functions of LLM layers vary with depth.
For example, \citet{liu2024fantastic} showed that the lower layers of LLaMA 2 encode lexical semantics, whereas the higher layers are responsible for next-token prediction.
This depth dependence leads us to hypothesize that lexical diversity in LLM outputs is more closely associated with lower layers, whereas semantic diversity is more closely associated with higher layers.
Under this hypothesis, it is expected that by utilizing the higher layers extracted from the base model, its semantic diversity can be brought into the merged model.
In addition, varying the split position is expected to adjust the balance between diversity and quality.

\section{Experiments}\label{sec:experiment}
\subsection{Diversity in Generated Text}\label{sec:experiment-diversity}
\subsubsection{Setup}
In this experiment, we evaluated how the output diversity of the merged models changes with the hyperparameter $i$ relative to that of the original models.

\paragraph{Preliminary}
This experiment evaluates per-input diversity: the diversity among $M$ outputs $y_1,\dots,y_{M}$ generated for a single input $x$, as opposed to across-input diversity (one output per input across different inputs).
We focus on per-input diversity because it is more directly relevant to downstream tasks such as data augmentation and story generation.

\paragraph{Evaluation metric}
We measured lexical diversity with Self-BLEU~\cite{zhu2018texygen} and semantic diversity with the cosine distance between sentence embeddings~\cite{reimers2019sentencebert}.
Self-BLEU evaluates each output with BLEU~\cite{papineni2002bleu} against the other outputs for the same input and averages the scores, so lower values indicate higher lexical diversity.
Let $\text{BLEU}(y;\mathcal{R})$ denote the BLEU score of text $y$ against a set of reference texts $\mathcal{R}$.
In this experiment, we defined a metric $\text{Div}_\text{lex}(y_1,\dots,y_M)$ as one minus Self-BLEU so that higher values indicate higher lexical diversity. Equation~\ref{eq:div_lex} gives this definition.
\begin{equation}
	\begin{split}
		 & \text{Div}_\text{lex}(y_1,\dots,y_M)                                                        \\
		 & \quad = 1-\frac{1}{M}\sum_{m=1}^{M}\text{BLEU}\left(y_m;\left\{y_k\right\}_{k\neq m}\right)
	\end{split}\label{eq:div_lex}
\end{equation}
The cosine distance between sentence embeddings is used to measure the semantic diversity of text.
Let $e(y)$ denote the sentence embedding of text $y$. Appendix~\ref{sec:appendix-settings} specifies the embedding model.
We defined a semantic diversity metric $\text{Div}_\text{sem}(y_1,\dots,y_M)$ as in Equation \ref{eq:div_sem}.
\begin{equation}
	\begin{split}
		 & \text{Div}_\text{sem}(y_1,\dots,y_M)                                                                                                                                                   \\
		 & \quad = \frac{1}{\binom{M}{2}}\sum_{1\leq m < k \leq M}\left(1-\frac{e\left(y_m\right)\cdot e\left(y_k\right)}{\left| e\left(y_m\right)\right| \left| e\left(y_k\right)\right|}\right)
	\end{split}\label{eq:div_sem}
\end{equation}
In this experiment, we set $M=10$.
That is, for each input $x$, we obtained 10 outputs $y_1,\dots,y_M$ and calculated $\text{Div}_\text{lex}$ and $\text{Div}_\text{sem}$ over these outputs.

\paragraph{Task}
This experiment addressed four tasks: translation, summarization, story generation, and dialogue generation.
For translation, we used the English--Japanese data of the WMT24++~\cite{deutsch2025wmt24} dataset in the Japanese-to-English direction: the Japanese texts served as inputs, and the original English texts served as references.
For summarization, we employed the XL-Sum~\cite{hasan2021xlsum} dataset, where the task involves generating summaries from input articles.
For story generation, we utilized the WritingPrompts~\cite{fan2018hierarchical} dataset, where the task involves generating a story from a writing prompt.
For dialogue generation, we utilized the DailyDialog~\cite{li2017dailydialog} dataset, where the task involves completing the subsequent utterances of a partial dialogue.
For each task, we used 500 randomly selected instruction examples $x$ and evaluated each model by the average $\text{Div}_\text{lex}$ and $\text{Div}_\text{sem}$ scores across all inputs.

\paragraph{Model}
We employed two models: LLaMA 2~\cite{touvron2023llamaa} and Qwen 2.5~\cite{qwen2024qwen2}.
For both models, we used versions with approximately 7B parameters: one that underwent only pretraining and another that was aligned (Appendix~\ref{sec:appendix-models}).
The value of $L$ was 32 for LLaMA 2 and 28 for Qwen 2.5.

\paragraph{Hyperparameter}
As a default hyperparameter, we set the temperature parameter to 1.0.
Furthermore, our approach has a hyperparameter $i$, and we conducted experiments across all values $i=1,\dots,L-1$.
Appendix~\ref{sec:appendix-settings} provides further details of the experimental settings, such as the prompts and the generation parameters.

\subsubsection{Result and Discussion}
\begin{figure*}[t]
	\centering
	\includegraphics[scale=0.7,trim={25pt 8pt 131pt 8pt},clip]{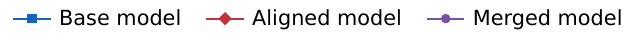}\\[1mm]
	\includegraphics[width=0.245\textwidth]{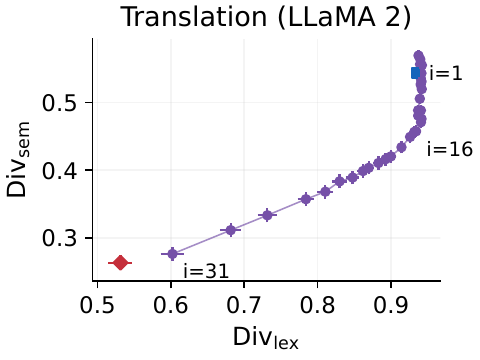}\hfill
	\includegraphics[width=0.245\textwidth]{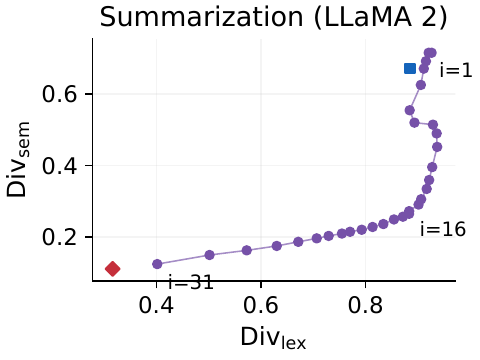}\hfill
	\includegraphics[width=0.245\textwidth]{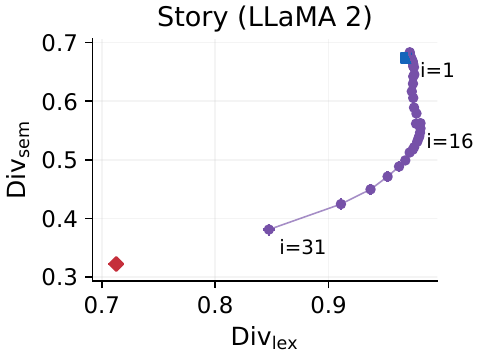}\hfill
	\includegraphics[width=0.245\textwidth]{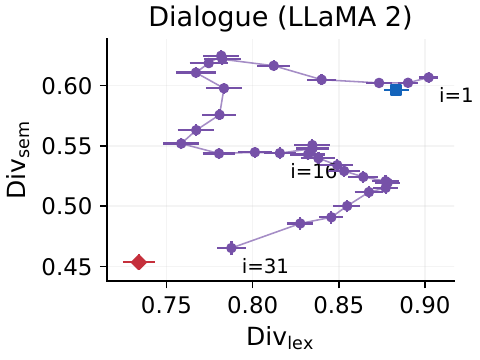}\\[2mm]
	\includegraphics[width=0.245\textwidth]{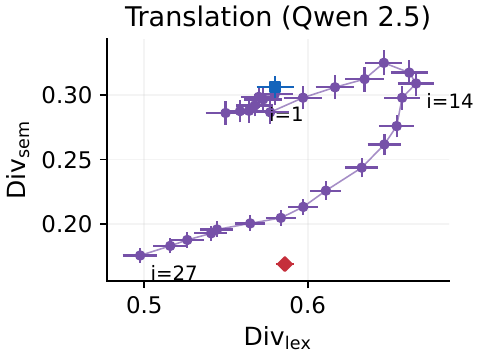}\hfill
	\includegraphics[width=0.245\textwidth]{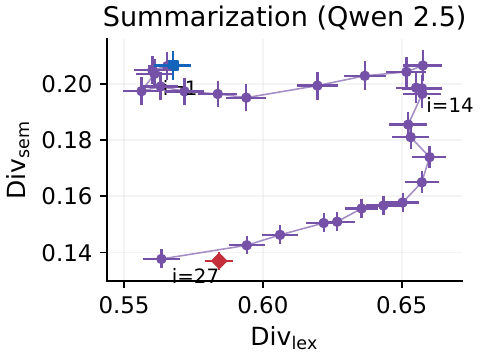}\hfill
	\includegraphics[width=0.245\textwidth]{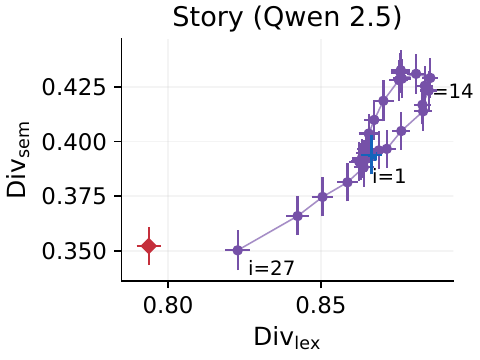}\hfill
	\includegraphics[width=0.245\textwidth]{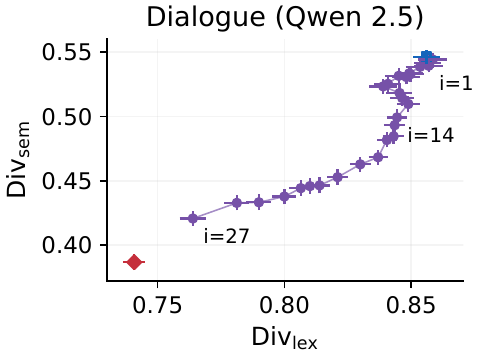}\\
	\caption{Lexical and semantic diversity of the base models (blue squares), the aligned models (red diamonds), and the merged models (purple circles) for LLaMA 2 (top row) and Qwen 2.5 (bottom row). The circles are connected in the order of $i$, and the labels indicate some values of $i$. The $x$-axis and $y$-axis denote $\text{Div}_\text{lex}$ and $\text{Div}_\text{sem}$, respectively. Each point is the average over the 500 inputs, and the error bars indicate 95\% confidence intervals obtained by 10,000 bootstrap resamples of the inputs.}
	\label{fig:diversity-llama2}
\end{figure*}

For each task, we plotted the two diversity metrics $\text{Div}_\text{lex}$ and $\text{Div}_\text{sem}$ for the base model, the aligned model, and the merged models for each value of the hyperparameter $i$.
The results are shown in Figure~\ref{fig:diversity-llama2}.

As $i$ increased, the semantic diversity of the merged models generally moved from that of the base model toward that of the aligned model.
This behavior is consistent with the layer composition of our approach, in which layers $1$ through $i$ are taken from the aligned model and layers $i+1$ through $L$ from the base model.
Smaller values of $i$ therefore correspond to a greater contribution from the base model, and larger values to a greater contribution from the aligned model, which accounts for the observed shift in semantic diversity.
In Qwen 2.5 story generation, however, the semantic diversity exceeded that of the base model at intermediate values of $i$.

The merged models also exhibited higher lexical diversity than both original models for some values of $i$ in many tasks.
The number of such values ranged from 2 in LLaMA 2 dialogue generation to 25 in LLaMA 2 story generation, and these values were mainly small to intermediate ones.
Furthermore, the merged models exhibited higher semantic diversity than the aligned model for almost all values of $i$.

\subsection{Quality of Generated Text}\label{sec:experiment-quality}
\subsubsection{Setup}
We evaluated how the output quality of the merged models changes with the hyperparameter $i$ relative to that of the original models.
We evaluated quality together with diversity because higher semantic diversity is not always desirable.
In translation and summarization, for example, outputs should remain faithful to the source text.
For this reason, we regard our approach as a means of controlling the trade-off between diversity and quality rather than maximizing diversity.

We focused on translation and summarization tasks, for which reference-based metrics are commonly used to measure output quality.
Although the datasets for story generation and dialogue generation also contain reference texts, these tasks allow many appropriate outputs for a single input, so similarity to a single reference does not necessarily indicate high quality.
We used the same datasets, number of inputs (500), models, and hyperparameters as in Section~\ref{sec:experiment-diversity}.
For evaluation metrics, we used COMET~\cite{rei2020comet} for translation and BERTScore~\cite{zhang2020bertscore} for summarization.
We averaged quality scores over 10 outputs per input and across all inputs.

\subsubsection{Result and Discussion}
\begin{figure}[tbp]
	\centering
	\includegraphics[scale=0.68,trim={15pt 8pt 19pt 8pt},clip]{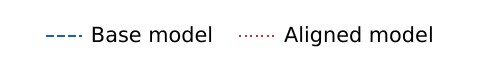}\hspace{0.6em}%
	\includegraphics[scale=0.68,trim={224pt 8pt 131pt 8pt},clip]{img/legends/confidence-legend.pdf}\\[1mm]
	\includegraphics[width=0.49\columnwidth]{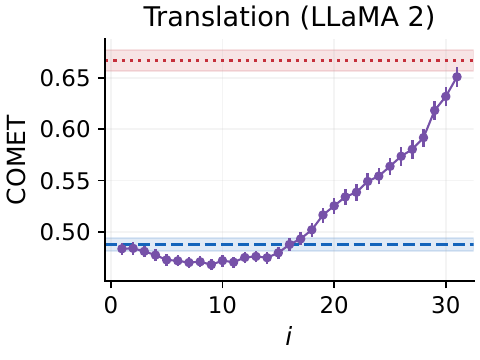}\hfill
	\includegraphics[width=0.49\columnwidth]{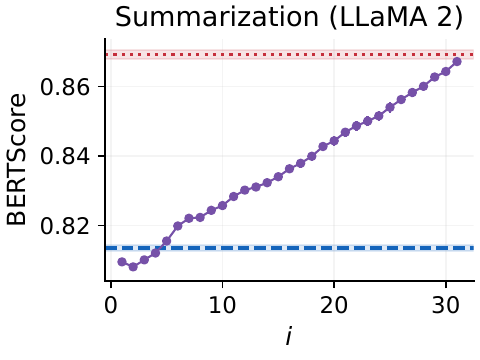}\\[2mm]
	\includegraphics[width=0.49\columnwidth]{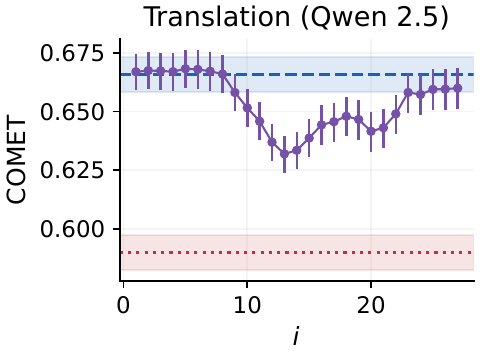}\hfill
	\includegraphics[width=0.49\columnwidth]{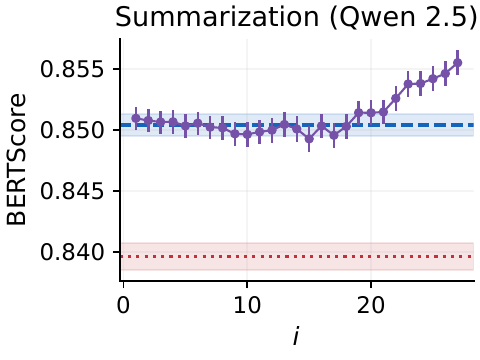}\\
	\caption{Output quality of the base models (blue dashed lines), the aligned models (red dotted lines), and the merged models (purple circles) for LLaMA 2 (top) and Qwen 2.5 (bottom). The $x$-axis denotes $i$, and the $y$-axis denotes COMET for translation (left) and BERTScore for summarization (right). Each point is the average over the 500 inputs, and the error bars and shaded bands indicate 95\% confidence intervals obtained by 10,000 bootstrap resamples of the inputs.}
	\label{fig:quality-llama2}
\end{figure}

For each task, we plotted quality evaluation metrics for the base model, the aligned model, and the merged models for each value of the hyperparameter $i$.
The results are shown in Figure~\ref{fig:quality-llama2}.

The output quality of the merged models changed with $i$ differently in the two model families.
In LLaMA 2, the quality of the merged models generally increased with $i$ and approached that of the aligned model at large values of $i$.
For Qwen 2.5, the aligned model performed worse than the base model in both tasks.
This is likely because all models received plain-text prompts in the input format of base models, without chat templates (Appendix~\ref{sec:appendix-prompts}).
Even so, the merged models showed higher quality than the aligned model for all values of $i$.
These results indicate that the hyperparameter $i$ adjusts the output quality of the merged models.

\begin{figure*}[t]
	\centering
	\includegraphics[scale=0.7,trim={25pt 8pt 131pt 8pt},clip]{img/legends/confidence-legend.pdf}\\[1mm]
	\includegraphics[width=0.245\textwidth]{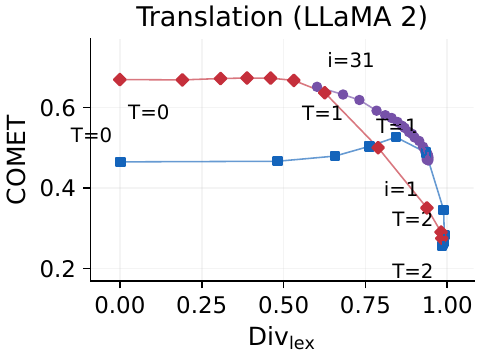}\hfill
	\includegraphics[width=0.245\textwidth]{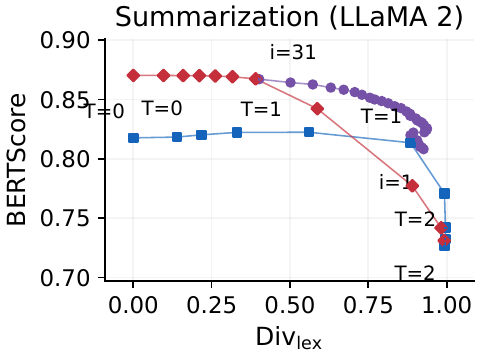}\hfill
	\includegraphics[width=0.245\textwidth]{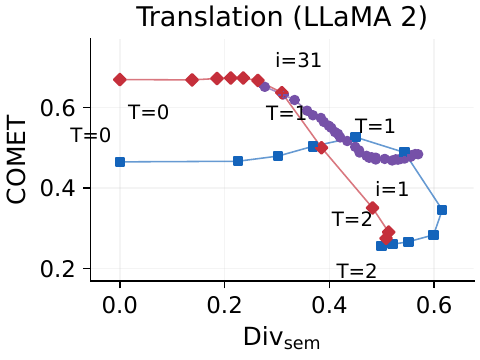}\hfill
	\includegraphics[width=0.245\textwidth]{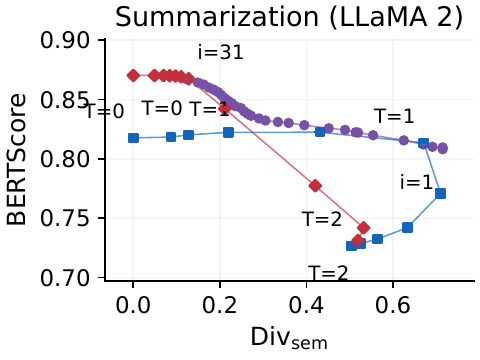}\\[2mm]
	\includegraphics[width=0.245\textwidth]{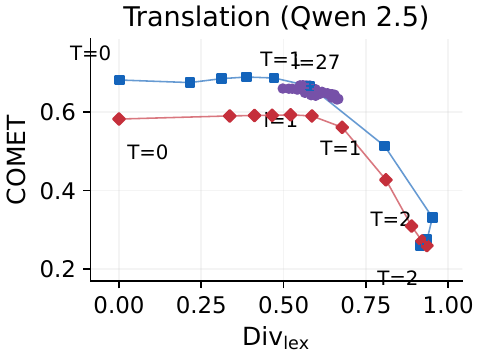}\hfill
	\includegraphics[width=0.245\textwidth]{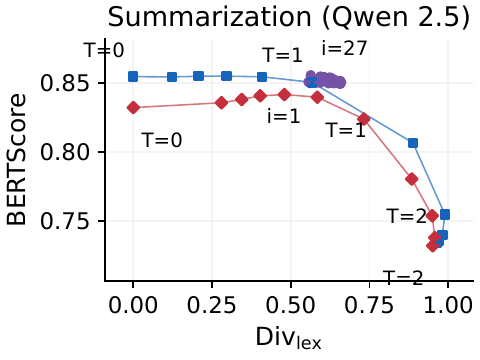}\hfill
	\includegraphics[width=0.245\textwidth]{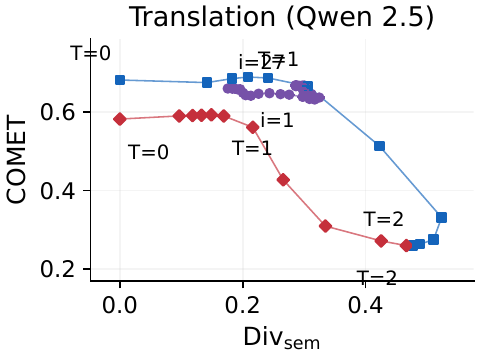}\hfill
	\includegraphics[width=0.245\textwidth]{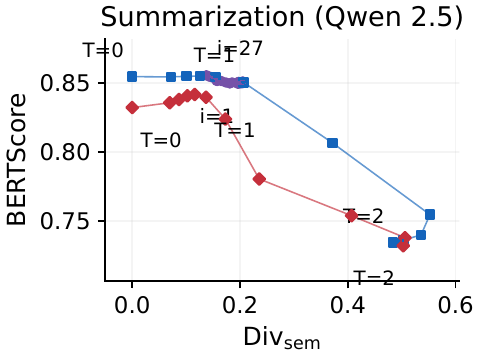}\\
	\caption{Trade-off between diversity (horizontal) and quality (vertical) for temperature sampling with the base models (blue squares) and the aligned models (red diamonds) and for the merged models (purple circles) for LLaMA 2 (top row) and Qwen 2.5 (bottom row). From left to right, the panels in each row show translation and summarization with $\text{Div}_\text{lex}$ and translation and summarization with $\text{Div}_\text{sem}$. The labels indicate some temperatures $T$ and values of $i$. Each point is the average over the 500 inputs, and the error bars indicate 95\% confidence intervals obtained by 10,000 bootstrap resamples of the inputs.}
	\label{fig:tradeoff-llama2}
\end{figure*}

\begin{table*}[t]
	\centering
	\small
	\setlength{\tabcolsep}{5pt}
	\begin{tabular}{llrrrll}
		\toprule
		Model    & Task          & Merged  & Base    & Aligned & Merged $-$ Base                     & Merged $-$ Aligned                  \\
		\midrule
		\multicolumn{7}{l}{\textit{Lexical diversity}}                                                                                     \\
		LLaMA 2  & Translation   & $43.80$ & $11.89$ & $15.98$ & $+31.91$ [$+29.80$, $+34.02$]$^{*}$ & $+27.82$ [$+26.16$, $+29.46$]$^{*}$ \\
		LLaMA 2  & Summarization & $25.99$ & $2.16$  & $9.64$  & $+23.83$ [$+23.18$, $+24.47$]$^{*}$ & $+16.35$ [$+15.89$, $+16.83$]$^{*}$ \\
		Qwen 2.5 & Translation   & $5.03$  & $0.00$  & $0.00$  & $+5.03$ [$+4.41$, $+5.71$]$^{*}$    & $+5.03$ [$+4.52$, $+5.71$]$^{*}$    \\
		Qwen 2.5 & Summarization & $1.27$  & $0.00$  & $0.00$  & $+1.27$ [$+1.19$, $+1.37$]$^{*}$    & $+1.27$ [$+1.19$, $+1.37$]$^{*}$    \\
		\midrule
		\multicolumn{7}{l}{\textit{Semantic diversity}}                                                                                    \\
		LLaMA 2  & Translation   & $17.61$ & $7.12$  & $7.71$  & $+10.50$ [$+9.53$, $+11.46$]$^{*}$  & $+9.91$ [$+9.04$, $+10.76$]$^{*}$   \\
		LLaMA 2  & Summarization & $10.34$ & $2.83$  & $3.34$  & $+7.51$ [$+7.15$, $+7.84$]$^{*}$    & $+7.01$ [$+6.63$, $+7.39$]$^{*}$    \\
		Qwen 2.5 & Translation   & $11.53$ & $11.99$ & $0.00$  & $-0.45$ [$-0.88$, $+0.21$]          & $+11.53$ [$+10.89$, $+12.42$]$^{*}$ \\
		Qwen 2.5 & Summarization & $0.85$  & $0.82$  & $0.00$  & $+0.03$ [$-0.00$, $+0.08$]          & $+0.85$ [$+0.79$, $+0.92$]$^{*}$    \\
		\bottomrule
	\end{tabular}
	\caption{Hypervolume ($\times 10^{-3}$) of the diversity--quality points of the merged models (all values of $i$) and of temperature sampling with the base and aligned models (all temperatures). The last two columns show the differences with 95\% confidence intervals obtained by a paired bootstrap over the 500 inputs (10,000 resamples). An asterisk ($^{*}$) indicates $p<0.05$ after Holm's correction.}
	\label{tab:hypervolume}
\end{table*}

\subsection{Comparison with Existing Method}\label{sec:experiment-comparison}
\subsubsection{Setup}
We compared the diversity--quality trade-off of our approach with that of temperature sampling.
We computed diversity and quality using the same tasks (translation, summarization), datasets, number of inputs (500), models, and hyperparameters as in Sections~\ref{sec:experiment-diversity} and \ref{sec:experiment-quality}.
We varied the temperature $T \in \{0.0, 0.2, \ldots, 2.0\}$ for temperature sampling and hyperparameter $i$ for our approach, then plotted the trade-offs.

We quantified each trade-off with the Hypervolume~\cite{zitzler1998multiobjective}.
The Hypervolume of a set of points is the area of the region that is dominated by at least one of the points and that dominates a reference point. A larger Hypervolume indicates a better set of trade-offs.
For the merged models, the set consists of the points for all values of $i$. For temperature sampling, it consists of the points for all temperatures.
We set the reference point to the component-wise minimum of the points of the base model and the aligned model at a temperature of 1.0.
We estimated 95\% confidence intervals of the differences in Hypervolume with a paired bootstrap over the 500 inputs (10,000 resamples) and recomputed the reference point in each resample. We tested the differences with two-sided bootstrap tests with Holm's correction for multiple comparisons.

\subsubsection{Result and Discussion}
\begin{figure*}[t]
	\centering
	\includegraphics[scale=0.7,trim={0 8pt 0 8pt},clip]{img/legends/confidence-legend.pdf}\\[1mm]
	\includegraphics[width=0.245\textwidth]{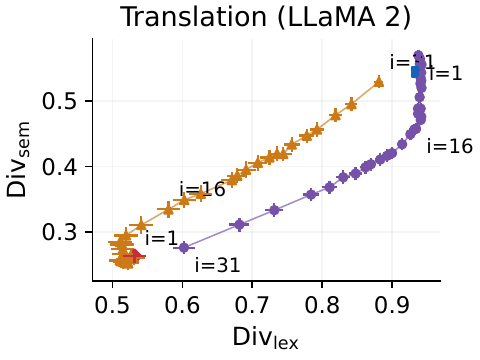}\hfill
	\includegraphics[width=0.245\textwidth]{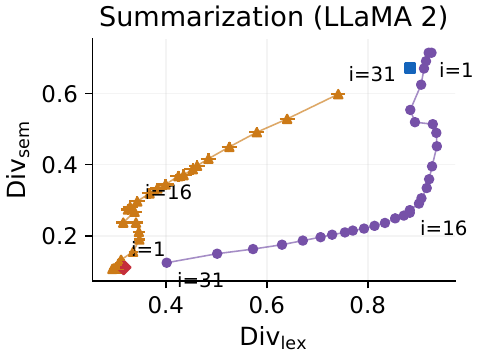}\hfill
	\includegraphics[width=0.245\textwidth]{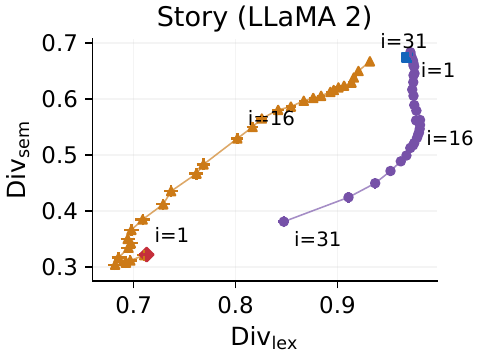}\hfill
	\includegraphics[width=0.245\textwidth]{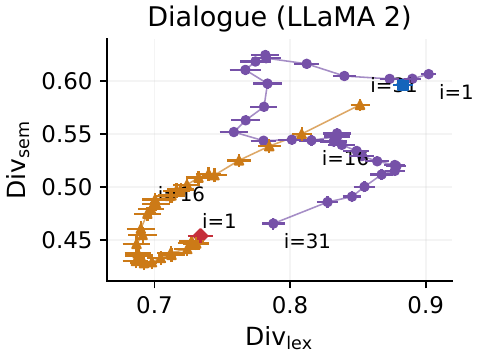}\\[2mm]
	\includegraphics[width=0.245\textwidth]{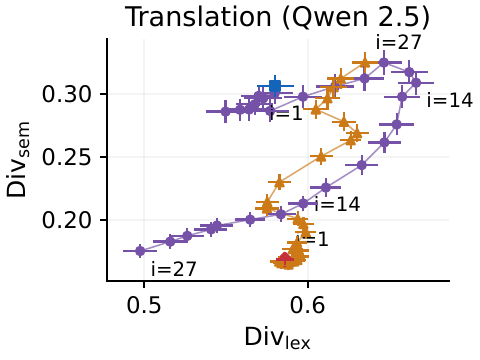}\hfill
	\includegraphics[width=0.245\textwidth]{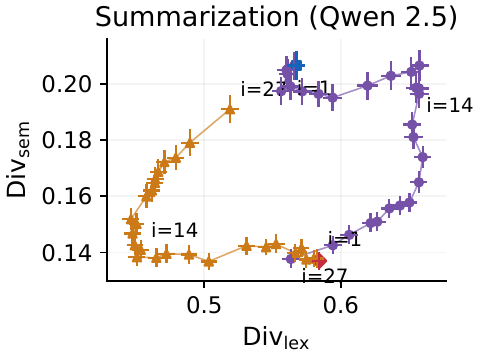}\hfill
	\includegraphics[width=0.245\textwidth]{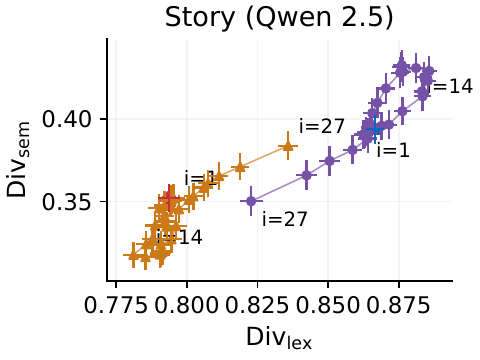}\hfill
	\includegraphics[width=0.245\textwidth]{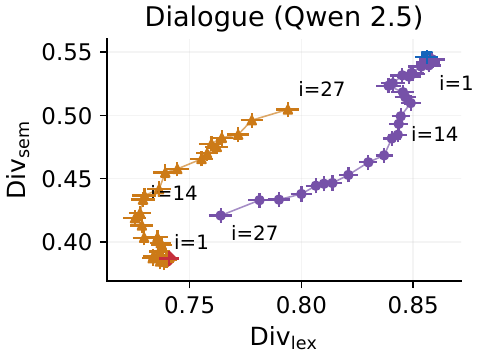}\\
	\caption{Diversity of the base models (blue squares), the aligned models (red diamonds), the merged models of our approach (purple circles), and the swapped models (orange triangles, labeled ``Inverse model'' in the legend) for LLaMA 2 (top row) and Qwen 2.5 (bottom row). The points of each merging order are connected in the order of $i$, and the labels indicate some values of $i$. The $x$-axis and $y$-axis denote $\text{Div}_\text{lex}$ and $\text{Div}_\text{sem}$, respectively. Each point is the average over the 500 inputs, and the error bars indicate 95\% confidence intervals obtained by 10,000 bootstrap resamples of the inputs.}
	\label{fig:diversity-llama2-ablation}
\end{figure*}

Figure \ref{fig:tradeoff-llama2} shows the trade-offs between diversity and quality metrics for the base model and the aligned model with temperature sampling, as well as for the merged models for each value of the hyperparameter $i$.
Table~\ref{tab:hypervolume} lists the Hypervolume of each method.

For both LLaMA 2 and Qwen 2.5, raising the temperature generally increased diversity on both tasks, whereas quality remained relatively stable up to a temperature of 1.0 and decreased sharply above it.
This is consistent with the findings of \citet{zhang2021trading} that temperature sampling enhances diversity at the cost of quality.

The Hypervolume of the merged models was larger than that of temperature sampling in most settings.
With lexical diversity, the Hypervolume of the merged models was significantly larger than those of temperature sampling with both the base model and the aligned model for both model families and both tasks.
With semantic diversity, the same held for LLaMA 2. For Qwen 2.5, the merged models significantly exceeded temperature sampling with the aligned model and were comparable to that with the base model.
In addition, in the plane of $\text{Div}_\text{lex}$ and quality, 30 or 31 of the 31 merged models for LLaMA 2 and 21 or 26 of the 27 merged models for Qwen 2.5 were not dominated by any point of temperature sampling.
In other words, no temperature yielded both diversity and quality at least as high as those of these merged models.
These results suggest that our approach can provide a better diversity--quality trade-off than temperature sampling, particularly in terms of lexical diversity.
We also observed that adjusting the hyperparameter $i$ enabled flexible control over the balance between the two.

\subsection{Ablation Study}\label{sec:experiment-ablation}
\subsubsection{Setup}
To examine the hypothesis that lexical diversity is more closely associated with lower layers and semantic diversity with higher layers, we conducted a supporting ablation.
We evaluated how output diversity changes when the merging order is reversed in our approach.
Specifically, we combined layers $1$ to $i$ from the base model with layers $i+1$ to $L$ from the aligned model.
We used the same metrics, models, tasks, and hyperparameters as in Section~\ref{sec:experiment-diversity}.

\subsubsection{Result and Discussion}
For each task, we plotted the two diversity metrics $\text{Div}_\text{lex}$ and $\text{Div}_\text{sem}$ for four model configurations: the base model, the aligned model, the models merged by our approach, and the swapped models.
In the swapped models, the order of the base model and the aligned model in our approach is reversed.
For the last two configurations, we plotted the results for each value of the hyperparameter $i$.
The results are shown in Figure \ref{fig:diversity-llama2-ablation}.

For both LLaMA 2 and Qwen 2.5, as $i$ increased, the semantic diversity of the swapped models approached that of the base model.
As in Section \ref{sec:experiment-diversity}, this suggests that the more layers a model inherits from a given source, the more strongly it is influenced by that source.

The swapped models showed lower lexical diversity than the merged models of our approach at most values of $i$.
To compare the two merging orders at the same value of $i$, we computed the difference in $\text{Div}_\text{lex}$ between them and its 95\% confidence interval with a paired bootstrap over the 500 inputs (10,000 resamples).
We adjusted the intervals for the multiple comparisons over the values of $i$ and the metrics and regarded the $\text{Div}_\text{lex}$ of our approach as significantly higher when the interval was above zero.
The $\text{Div}_\text{lex}$ of our approach was significantly higher than that of the swapped models for 27 to 30 of the 31 values of $i$ in LLaMA 2, for 22 to 26 of the 27 values in Qwen 2.5 summarization, story generation, and dialogue generation, and for 9 of the 27 values in Qwen 2.5 translation.
Moreover, for some values of $i$, the lexical diversity of the swapped models fell below even that of the aligned model in all settings.
This suggests that the higher lexical diversity of our approach may be related to its use of the lower layers of the aligned model.
On the other hand, the semantic diversity of the swapped models remained at or below that of the base model for almost all values of $i$.
This suggests that the higher semantic diversity of our approach may be related to its use of the higher layers of the base model.
Overall, these results are largely consistent with the hypothesis that lexical diversity is more closely associated with lower layers and semantic diversity with higher layers, and may reflect a depth-dependent effect on output diversity.

\section{Conclusion}\label{sec:conclusion}
In this study, we introduce a depth-aware FrankenMerging approach that combines the lower layers of an aligned LLM with the higher layers of its base model to control the diversity--quality trade-off between the two models.
Experiments on four tasks showed that the number of layers taken from each model allows us to select an operating point on this trade-off according to the requirements of each task.
The merged models also showed a better diversity--quality trade-off than temperature sampling in most settings, particularly in terms of lexical diversity.
Furthermore, the results of the supporting ablation in which we reversed the merging order were largely consistent with the hypothesis that lexical diversity is more closely associated with lower layers and semantic diversity with higher layers, which may reflect a depth-dependent effect on output diversity.

\section*{Limitations}
Our experiments focus on two model families with approximately 7B parameters, LLaMA 2 and Qwen 2.5, which were aligned with different procedures.
Extending the experiments to other model families and scales may provide further insights.
In addition, since the trade-off range between quality and diversity that our approach can control is limited to that of the base model and the aligned model, the quality and diversity of the merged model depend on those of these two models.
This limitation is expected to be resolved as the performance of base models improves.

Our analysis of layer depth compares the two merging orders.
Finer-grained analyses, for example the replacement of a fixed number of layers at different depths, would provide further insight into the roles of individual layers.

The appropriate value of $i$ depends on the task and the purpose of use.
For practical applications such as data augmentation, we suggest adjusting $i$ automatically, for example with validation data of the target task.
Evaluating such adjustment in real applications is future work.

Our approach modifies the model rather than the decoding procedure, so it can be combined with sampling-based methods and diversity-oriented fine-tuning.
Such combinations are expected to further improve output diversity.

\section*{Ethical Considerations}
We used the models and datasets in accordance with their licenses and terms of use.
LLaMA 2 is released under the Llama 2 Community License, and Qwen 2.5 and Qwen3-Embedding are released under the Apache License 2.0.
Of the datasets, WMT24++ is released under the Apache License 2.0, XL-Sum and DailyDialog under CC BY-NC-SA 4.0, and the version of WritingPrompts that we used under the MIT License.
We used all of them only for research purposes, which is consistent with these licenses.

The merged models may not fully retain the safety behavior that the aligned models acquired through alignment.
This is because our approach takes the higher layers of a merged model from its base model.
Before using merged models in applications, practitioners should evaluate their safety in addition to their diversity and quality.

WritingPrompts consists of prompts and stories posted on Reddit and may contain offensive content.
We used these data only as inputs for generation and did not release the generated outputs.

\bibliography{main}

@article{ackley1985learning,
  title = {{A Learning Algorithm for {Boltzmann} Machines}},
  author = {Ackley, David H. and Hinton, Geoffrey E. and Sejnowski, Terrence J.},
  year = 1985,
  month = jan,
  journal = {Cognitive Science},
  volume = {9},
  number = {1},
  pages = {147--169},
  issn = {0364-0213},
  doi = {10.1016/S0364-0213(85)80012-4}
}

@inproceedings{ainsworth2023git,
  title = {{Git Re-Basin: Merging Models Modulo Permutation Symmetries}},
  shorttitle = {Git {{Re-Basin}}},
  booktitle = {The {{Eleventh International Conference}} on {{Learning Representations}}},
  author = {Ainsworth, Samuel and Hayase, Jonathan and Srinivasa, Siddhartha},
  year = 2023
}

@article{akiba2025evolutionary,
  title = {{Evolutionary Optimization of Model Merging Recipes}},
  author = {Akiba, Takuya and Shing, Makoto and Tang, Yujin and Sun, Qi and Ha, David},
  year = 2025,
  month = feb,
  journal = {Nature Machine Intelligence},
  volume = {7},
  number = {2},
  pages = {195--204},
  issn = {2522-5839},
  doi = {10.1038/s42256-024-00975-8}
}

@inproceedings{chung2025modifying,
  title = {{Modifying Large Language Model Post-Training for Diverse Creative Writing}},
  author={John Joon Young Chung and Vishakh Padmakumar and Melissa Roemmele and Yuqian Sun and Max Kreminski},
  booktitle={Second Conference on Language Modeling},
  year={2025},
  url={https://openreview.net/forum?id=1Pmuw08LoM}
}

@inproceedings{deutsch2025wmt24,
  title = {{WMT24++: Expanding the Language Coverage of WMT24 to 55 Languages \& Dialects}},
  shorttitle = {{{WMT24}}++},
  booktitle = {Findings of the {{Association}} for {{Computational Linguistics}}: {{ACL}} 2025},
  author = {Deutsch, Daniel and Briakou, Eleftheria and Caswell, Isaac Rayburn and Finkelstein, Mara and Galor, Rebecca and Juraska, Juraj and Kovacs, Geza and Lui, Alison and Rei, Ricardo and Riesa, Jason and Rijhwani, Shruti and Riley, Parker and Salesky, Elizabeth and Trabelsi, Firas and Winkler, Stephanie and Zhang, Biao and Freitag, Markus},
  editor = {Che, Wanxiang and Nabende, Joyce and Shutova, Ekaterina and Pilehvar, Mohammad Taher},
  year = 2025,
  month = jul,
  pages = {12257--12284},
  publisher = {Association for Computational Linguistics},
  address = {Vienna, Austria},
  doi = {10.18653/v1/2025.findings-acl.634},
  isbn = {979-8-89176-256-5}
}

@inproceedings{fan2018hierarchical,
  title = {{Hierarchical Neural Story Generation}},
  booktitle = {Proceedings of the 56th {{Annual Meeting}} of the {{Association}} for {{Computational Linguistics}} ({{Volume}} 1: {{Long Papers}})},
  author = {Fan, Angela and Lewis, Mike and Dauphin, Yann},
  editor = {Gurevych, Iryna and Miyao, Yusuke},
  year = 2018,
  month = jul,
  pages = {889--898},
  publisher = {Association for Computational Linguistics},
  address = {Melbourne, Australia},
  doi = {10.18653/v1/P18-1082}
}

@inproceedings{goddard2024arcees,
  title = {{Arcee's MergeKit: A Toolkit for Merging Large Language Models}},
  shorttitle = {Arcee's {{MergeKit}}},
  booktitle = {Proceedings of the 2024 {{Conference}} on {{Empirical Methods}} in {{Natural Language Processing}}: {{Industry Track}}},
  author = {Goddard, Charles and Siriwardhana, Shamane and Ehghaghi, Malikeh and Meyers, Luke and Karpukhin, Vladimir and Benedict, Brian and McQuade, Mark and Solawetz, Jacob},
  editor = {Dernoncourt, Franck and {Preo{\c t}iuc-Pietro}, Daniel and Shimorina, Anastasia},
  year = 2024,
  month = jan,
  pages = {477--485},
  publisher = {Association for Computational Linguistics},
  address = {Miami, Florida, US},
  doi = {10.18653/v1/2024.emnlp-industry.36}
}

@inproceedings{guo2024curious,
  title = {{The Curious Decline of Linguistic Diversity: Training Language Models on Synthetic Text}},
  shorttitle = {The {{Curious Decline}} of {{Linguistic Diversity}}},
  booktitle = {Findings of the {{Association}} for {{Computational Linguistics}}: {{NAACL}} 2024},
  author = {Guo, Yanzhu and Shang, Guokan and Vazirgiannis, Michalis and Clavel, Chlo{\'e}},
  editor = {Duh, Kevin and Gomez, Helena and Bethard, Steven},
  year = 2024,
  month = jun,
  pages = {3589--3604},
  publisher = {Association for Computational Linguistics},
  address = {Mexico City, Mexico},
  doi = {10.18653/v1/2024.findings-naacl.228}
}

@article{guo2025benchmarking,
  title = {{Benchmarking Linguistic Diversity of Large Language Models}},
  author = {Guo, Yanzhu and Shang, Guokan and Clavel, Chlo{\'e}},
  year = 2025,
  month = nov,
  journal = {Transactions of the Association for Computational Linguistics},
  volume = {13},
  pages = {1507--1526},
  issn = {2307-387X},
  doi = {10.1162/TACL.a.47}
}

@inproceedings{hasan2021xlsum,
  title = {{XL-Sum: Large-Scale Multilingual Abstractive Summarization for 44 Languages}},
  shorttitle = {{{XL-Sum}}},
  booktitle = {Findings of the {{Association}} for {{Computational Linguistics}}: {{ACL-IJCNLP}} 2021},
  author = {Hasan, Tahmid and Bhattacharjee, Abhik and Islam, Md. Saiful and Mubasshir, Kazi and Li, Yuan-Fang and Kang, Yong-Bin and Rahman, M. Sohel and Shahriyar, Rifat},
  editor = {Zong, Chengqing and Xia, Fei and Li, Wenjie and Navigli, Roberto},
  year = 2021,
  month = aug,
  pages = {4693--4703},
  publisher = {Association for Computational Linguistics},
  address = {Online},
  doi = {10.18653/v1/2021.findings-acl.413}
}

@inproceedings{kirk2024understanding,
  title = {{Understanding the Effects of RLHF on LLM Generalisation and Diversity}},
  booktitle = {The {{Twelfth International Conference}} on {{Learning Representations}}},
  author = {Kirk, Robert and Mediratta, Ishita and Nalmpantis, Christoforos and Luketina, Jelena and Hambro, Eric and Grefenstette, Edward and Raileanu, Roberta},
  year = 2024
}

@article{lanchantin2025diverse,
  title = {{Diverse Preference Optimization}},
  author={Lanchantin, Jack and Chen, Angelica and Dhuliawala, Shehzaad and Yu, Ping and Weston, Jason and Sukhbaatar, Sainbayar and Kulikov, Ilia},
  journal={arXiv:2501.18101v4 [cs.CL]},
  year={2025}
}

@inproceedings{li2016diversitypromoting,
  title = {{A Diversity-Promoting Objective Function for Neural Conversation Models}},
  booktitle = {Proceedings of the 2016 {{Conference}} of the {{North American Chapter}} of the {{Association}} for {{Computational Linguistics}}: {{Human Language Technologies}}},
  author = {Li, Jiwei and Galley, Michel and Brockett, Chris and Gao, Jianfeng and Dolan, Bill},
  editor = {Knight, Kevin and Nenkova, Ani and Rambow, Owen},
  year = 2016,
  month = jun,
  pages = {110--119},
  publisher = {Association for Computational Linguistics},
  address = {San Diego, California},
  doi = {10.18653/v1/N16-1014}
}

@inproceedings{li2017dailydialog,
  title = {{DailyDialog: A Manually Labelled Multi-Turn Dialogue Dataset}},
  shorttitle = {{{DailyDialog}}},
  booktitle = {Proceedings of the {{Eighth International Joint Conference}} on {{Natural Language Processing}} ({{Volume}} 1: {{Long Papers}})},
  author = {Li, Yanran and Su, Hui and Shen, Xiaoyu and Li, Wenjie and Cao, Ziqiang and Niu, Shuzi},
  editor = {Kondrak, Greg and Watanabe, Taro},
  year = 2017,
  month = jan,
  pages = {986--995},
  publisher = {Asian Federation of Natural Language Processing},
  address = {Taipei, Taiwan}
}

@inproceedings{liu2024fantastic,
  title = {{Fantastic Semantics and Where to Find Them: Investigating Which Layers of Generative LLMs Reflect Lexical Semantics}},
  shorttitle = {Fantastic {{Semantics}} and {{Where}} to {{Find Them}}},
  booktitle = {Findings of the {{Association}} for {{Computational Linguistics}}: {{ACL}} 2024},
  author = {Liu, Zhu and Kong, Cunliang and Liu, Ying and Sun, Maosong},
  editor = {Ku, Lun-Wei and Martins, Andre and Srikumar, Vivek},
  year = 2024,
  month = aug,
  pages = {14551--14558},
  publisher = {Association for Computational Linguistics},
  address = {Bangkok, Thailand},
  doi = {10.18653/v1/2024.findings-acl.866}
}

@article{meister2023locally,
  title = {{Locally Typical Sampling}},
  author = {Meister, Clara and Pimentel, Tiago and Wiher, Gian and Cotterell, Ryan},
  year = 2023,
  journal = {Transactions of the Association for Computational Linguistics},
  volume = {11},
  pages = {102--121},
  publisher = {MIT Press},
  address = {Cambridge, MA},
  doi = {10.1162/tacl_a_00536}
}

@inproceedings{minh2025turning,
  title = {{Turning Up the Heat: Min-p Sampling for Creative and Coherent LLM Outputs}},
  shorttitle = {Turning {{Up}} the {{Heat}}},
  booktitle = {The {{Thirteenth International Conference}} on {{Learning Representations}}},
  author = {Minh, Nguyen Nhat and Baker, Andrew and Neo, Clement and Roush, Allen G. and Kirsch, Andreas and {Shwartz-Ziv}, Ravid},
  year = 2025
}

@inproceedings{papineni2002bleu,
  title = {{Bleu: A Method for Automatic Evaluation of Machine Translation}},
  shorttitle = {Bleu},
  booktitle = {Proceedings of the 40th {{Annual Meeting}} of the {{Association}} for {{Computational Linguistics}}},
  author = {Papineni, Kishore and Roukos, Salim and Ward, Todd and Zhu, Wei-Jing},
  editor = {Isabelle, Pierre and Charniak, Eugene and Lin, Dekang},
  year = 2002,
  month = jul,
  pages = {311--318},
  publisher = {Association for Computational Linguistics},
  address = {Philadelphia, Pennsylvania, USA},
  doi = {10.3115/1073083.1073135}
}

@article{qwen2024qwen2,
  title = {{Qwen2.5 Technical Report}},
  author={Qwen and An Yang and Baosong Yang and Beichen Zhang and Binyuan Hui and Bo Zheng and Bowen Yu and Chengyuan Li and Dayiheng Liu and Fei Huang and Haoran Wei and Huan Lin and Jian Yang and Jianhong Tu and Jianwei Zhang and Jianxin Yang and Jiaxi Yang and Jingren Zhou and Junyang Lin and Kai Dang and Keming Lu and Keqin Bao and Kexin Yang and Le Yu and Mei Li and Mingfeng Xue and Pei Zhang and Qin Zhu and Rui Men and Runji Lin and Tianhao Li and Tianyi Tang and Tingyu Xia and Xingzhang Ren and Xuancheng Ren and Yang Fan and Yang Su and Yichang Zhang and Yu Wan and Yuqiong Liu and Zeyu Cui and Zhenru Zhang and Zihan Qiu},
  journal={arXiv:2412.15115v2 [cs.CL]},
  year={2024}
}

@article{rafailov2023direct,
  title = {{Direct Preference Optimization: Your Language Model Is Secretly a Reward Model}},
  shorttitle = {Direct {{Preference Optimization}}},
  author = {Rafailov, Rafael and Sharma, Archit and Mitchell, Eric and Manning, Christopher D. and Ermon, Stefano and Finn, Chelsea},
  year = 2023,
  month = dec,
  journal = {Advances in Neural Information Processing Systems},
  volume = {36},
  pages = {53728--53741}
}

@inproceedings{rei2020comet,
  title = {{COMET: A Neural Framework for MT Evaluation}},
  shorttitle = {{{COMET}}},
  booktitle = {Proceedings of the 2020 {{Conference}} on {{Empirical Methods}} in {{Natural Language Processing}} ({{EMNLP}})},
  author = {Rei, Ricardo and Stewart, Craig and Farinha, Ana C and Lavie, Alon},
  editor = {Webber, Bonnie and Cohn, Trevor and He, Yulan and Liu, Yang},
  year = 2020,
  month = jan,
  pages = {2685--2702},
  publisher = {Association for Computational Linguistics},
  address = {Online},
  doi = {10.18653/v1/2020.emnlp-main.213}
}

@inproceedings{reimers2019sentencebert,
  title = {{Sentence-BERT: Sentence Embeddings Using Siamese BERT-Networks}},
  shorttitle = {Sentence-{{BERT}}},
  booktitle = {Proceedings of the 2019 {{Conference}} on {{Empirical Methods}} in {{Natural Language Processing}} and the 9th {{International Joint Conference}} on {{Natural Language Processing}} ({{EMNLP-IJCNLP}})},
  author = {Reimers, Nils and Gurevych, Iryna},
  editor = {Inui, Kentaro and Jiang, Jing and Ng, Vincent and Wan, Xiaojun},
  year = 2019,
  month = jan,
  pages = {3982--3992},
  publisher = {Association for Computational Linguistics},
  address = {Hong Kong, China},
  doi = {10.18653/v1/D19-1410}
}

@inproceedings{shypula2025evaluating,
  title = {{Evaluating the Diversity and Quality of LLM Generated Content}},
  booktitle = {Second {{Conference}} on {{Language Modeling}}},
  author = {Shypula, Alexander and Li, Shuo and Zhang, Botong and Padmakumar, Vishakh and Yin, Kayo and Bastani, Osbert},
  year = 2025,
  month = aug
}

@inproceedings{stasaski2022semantic,
  title = {{Semantic Diversity in Dialogue with Natural Language Inference}},
  booktitle = {Proceedings of the 2022 {{Conference}} of the {{North American Chapter}} of the {{Association}} for {{Computational Linguistics}}: {{Human Language Technologies}}},
  author = {Stasaski, Katherine and Hearst, Marti},
  editor = {Carpuat, Marine and {de Marneffe}, Marie-Catherine and Meza Ruiz, Ivan Vladimir},
  year = 2022,
  month = jul,
  pages = {85--98},
  publisher = {Association for Computational Linguistics},
  address = {Seattle, United States},
  doi = {10.18653/v1/2022.naacl-main.6}
}

@inproceedings{tevet2021evaluating,
  title = {{Evaluating the Evaluation of Diversity in Natural Language Generation}},
  booktitle = {Proceedings of the 16th {{Conference}} of the {{European Chapter}} of the {{Association}} for {{Computational Linguistics}}: {{Main Volume}}},
  author = {Tevet, Guy and Berant, Jonathan},
  editor = {Merlo, Paola and Tiedemann, Jorg and Tsarfaty, Reut},
  year = 2021,
  month = apr,
  pages = {326--346},
  publisher = {Association for Computational Linguistics},
  address = {Online},
  doi = {10.18653/v1/2021.eacl-main.25}
}

@article{touvron2023llamaa,
  title = {{Llama 2: Open Foundation and Fine-Tuned Chat Models}},
  author={Hugo Touvron and Louis Martin and Kevin Stone and Peter Albert and Amjad Almahairi and Yasmine Babaei and Nikolay Bashlykov and Soumya Batra and Prajjwal Bhargava and Shruti Bhosale and Dan Bikel and Lukas Blecher and Cristian Canton Ferrer and Moya Chen and Guillem Cucurull and David Esiobu and Jude Fernandes and Jeremy Fu and Wenyin Fu and Brian Fuller and Cynthia Gao and Vedanuj Goswami and Naman Goyal and Anthony Hartshorn and Saghar Hosseini and Rui Hou and Hakan Inan and Marcin Kardas and Viktor Kerkez and Madian Khabsa and Isabel Kloumann and Artem Korenev and Punit Singh Koura and Marie-Anne Lachaux and Thibaut Lavril and Jenya Lee and Diana Liskovich and Yinghai Lu and Yuning Mao and Xavier Martinet and Todor Mihaylov and Pushkar Mishra and Igor Molybog and Yixin Nie and Andrew Poulton and Jeremy Reizenstein and Rashi Rungta and Kalyan Saladi and Alan Schelten and Ruan Silva and Eric Michael Smith and Ranjan Subramanian and Xiaoqing Ellen Tan and Binh Tang and Ross Taylor and Adina Williams and Jian Xiang Kuan and Puxin Xu and Zheng Yan and Iliyan Zarov and Yuchen Zhang and Angela Fan and Melanie Kambadur and Sharan Narang and Aurelien Rodriguez and Robert Stojnic and Sergey Edunov and Thomas Scialom},
  journal={arXiv:2307.09288v2 [cs.CL]},
  year={2023}
}

@inproceedings{wang2025diversityoriented,
  title = {{Diversity-Oriented Data Augmentation with Large Language Models}},
  booktitle = {Proceedings of the 63rd {{Annual Meeting}} of the {{Association}} for {{Computational Linguistics}} ({{Volume}} 1: {{Long Papers}})},
  author = {Wang, Zaitian and Zhang, Jinghan and Zhang, Xinhao and Liu, Kunpeng and Wang, Pengfei and Zhou, Yuanchun},
  editor = {Che, Wanxiang and Nabende, Joyce and Shutova, Ekaterina and Pilehvar, Mohammad Taher},
  year = 2025,
  month = jul,
  pages = {22265--22283},
  publisher = {Association for Computational Linguistics},
  address = {Vienna, Austria},
  doi = {10.18653/v1/2025.acl-long.1084},
  isbn = {979-8-89176-251-0}
}

@inproceedings{wortsman2022model,
  title = {{Model Soups: Averaging Weights of Multiple Fine-Tuned Models Improves Accuracy without Increasing Inference Time}},
  shorttitle = {Model Soups},
  booktitle = {Proceedings of the 39th {{International Conference}} on {{Machine Learning}}},
  author = {Wortsman, Mitchell and Ilharco, Gabriel and Gadre, Samir Ya and Roelofs, Rebecca and {Gontijo-Lopes}, Raphael and Morcos, Ari S. and Namkoong, Hongseok and Farhadi, Ali and Carmon, Yair and Kornblith, Simon and Schmidt, Ludwig},
  year = 2022,
  month = jun,
  pages = {23965--23998},
  publisher = {PMLR},
  issn = {2640-3498}
}

@article{xu2025echoes,
  title = {{Echoes in AI: Quantifying Lack of Plot Diversity in LLM Outputs}},
  shorttitle = {Echoes in {{AI}}},
  author = {Xu, Weijia and Jojic, Nebojsa and Rao, Sudha and Brockett, Chris and Dolan, Bill},
  year = 2025,
  month = sep,
  journal = {Proceedings of the National Academy of Sciences},
  volume = {122},
  number = {35},
  pages = {e2504966122},
  publisher = {Proceedings of the National Academy of Sciences},
  doi = {10.1073/pnas.2504966122}
}

@inproceedings{zhang2020bertscore,
  title = {{BERTScore: Evaluating Text Generation with BERT}},
  shorttitle = {{{BERTScore}}},
  booktitle = {International {{Conference}} on {{Learning Representations}}},
  author = {Zhang, Tianyi and Kishore, Varsha and Wu, Felix and Weinberger, Kilian Q. and Artzi, Yoav},
  year = 2020
}

@inproceedings{zhang2021trading,
  title = {{Trading Off Diversity and Quality in Natural Language Generation}},
  booktitle = {Proceedings of the {{Workshop}} on {{Human Evaluation}} of {{NLP Systems}} ({{HumEval}})},
  author = {Zhang, Hugh and Duckworth, Daniel and Ippolito, Daphne and Neelakantan, Arvind},
  editor = {Belz, Anya and Agarwal, Shubham and Graham, Yvette and Reiter, Ehud and Shimorina, Anastasia},
  year = 2021,
  month = apr,
  pages = {25--33},
  publisher = {Association for Computational Linguistics},
  address = {Online}
}

@inproceedings{zhu2018texygen,
  title = {{Texygen: A Benchmarking Platform for Text Generation Models}},
  booktitle = {The 41st {{International ACM SIGIR Conference}} on {{Research}} \& {{Development}} in {{Information Retrieval}}},
  author = {Zhu, Yaoming and Lu, Sidi and Zheng, Lei and Guo, Jiaxian and Zhang, Weinan and Wang, Jun and Yu, Yong},
  year = 2018,
  pages = {1097--1100},
  doi = {10.1145/3209978.3210080}
}

@inproceedings{zitzler1998multiobjective,
  title = {{Multiobjective Optimization Using Evolutionary Algorithms: A Comparative Case Study}},
  booktitle = {Parallel {{Problem Solving}} from {{Nature}} --- {{PPSN V}}},
  author = {Zitzler, Eckart and Thiele, Lothar},
  year = 1998,
  pages = {292--301},
  publisher = {Springer},
  doi = {10.1007/BFb0056872}
}

@inproceedings{rei2022comet22,
  title = {{COMET-22: Unbabel-IST 2022 Submission for the Metrics Shared Task}},
  author = {Rei, Ricardo and C. de Souza, Jos{\'e} G. and Alves, Duarte and Zerva, Chrysoula and Farinha, Ana C and Glushkova, Taisiya and Lavie, Alon and Coheur, Luisa and Martins, Andr{\'e} F. T.},
  booktitle = {Proceedings of the Seventh Conference on Machine Translation (WMT)},
  year = 2022,
  pages = {578--585},
  address = {Abu Dhabi, United Arab Emirates (Hybrid)},
  publisher = {Association for Computational Linguistics},
  url = {https://aclanthology.org/2022.wmt-1.52}
}

@article{zhang2025qwen3,
  title = {{Qwen3 Embedding: Advancing Text Embedding and Reranking Through Foundation Models}},
  author = {Zhang, Yanzhao and Li, Mingxin and Long, Dingkun and Zhang, Xin and Lin, Huan and Yang, Baosong and Xie, Pengjun and Yang, An and Liu, Dayiheng and Lin, Junyang and Huang, Fei and Zhou, Jingren},
  journal = {arXiv:2506.05176v3 [cs.CL]},
  year = {2025}
}

@inproceedings{kwon2023efficient,
  title = {{Efficient Memory Management for Large Language Model Serving with PagedAttention}},
  author = {Kwon, Woosuk and Li, Zhuohan and Zhuang, Siyuan and Sheng, Ying and Zheng, Lianmin and Yu, Cody Hao and Gonzalez, Joseph E. and Zhang, Hao and Stoica, Ion},
  booktitle = {Proceedings of the 29th Symposium on Operating Systems Principles},
  year = 2023,
  pages = {611--626},
  publisher = {Association for Computing Machinery},
  doi = {10.1145/3600006.3613165}
}

@article{liu2019roberta,
  title = {{RoBERTa: A Robustly Optimized BERT Pretraining Approach}},
  author = {Liu, Yinhan and Ott, Myle and Goyal, Naman and Du, Jingfei and Joshi, Mandar and Chen, Danqi and Levy, Omer and Lewis, Mike and Zettlemoyer, Luke and Stoyanov, Veselin},
  journal = {arXiv:1907.11692v1 [cs.CL]},
  year = {2019}
}

\appendix

\section{Experimental Settings}\label{sec:appendix-settings}
This appendix provides the settings that Section~\ref{sec:experiment} does not describe.

\subsection{Models}\label{sec:appendix-models}
% Left-aligned to avoid wide spaces around the long URLs.
{\raggedright
	We obtained all models from the Hugging Face Hub.
	For LLaMA 2, we used \url{https://huggingface.co/meta-llama/Llama-2-7b-hf} as the base model and \url{https://huggingface.co/meta-llama/Llama-2-7b-chat-hf} as the aligned model.
	For Qwen 2.5, we used \url{https://huggingface.co/Qwen/Qwen2.5-7B} as the base model and \url{https://huggingface.co/Qwen/Qwen2.5-7B-Instruct} as the aligned model.\par}

\subsection{Merging}
We merged the models with the \texttt{passthrough} method of MergeKit~\cite{goddard2024arcees}, which applies no weight averaging, rescaling, or additional training.
Figure~\ref{fig:appendix-mergekit} shows the configuration.
MergeKit takes the input embeddings from the model of the first slice and takes the final normalization layer and the language modeling head from the model of the second slice.
The merged models keep the data types of the original weights, which are float16 for LLaMA 2 and bfloat16 for Qwen 2.5.
Their model configurations follow those of the base models, except for the maximum context length and the EOS tokens in Table~\ref{tab:appendix-generation}.
All conditions use the tokenizer of the base model, which has the same vocabulary as that of the aligned model.

\begin{figure}[t]
	\small
	\begin{verbatim}
merge_method: passthrough
base_model: <base model>
slices:
  - sources:
      - model: <aligned model>
        layer_range: [0, i]
  - sources:
      - model: <base model>
        layer_range: [i, L]
\end{verbatim}
	\caption{MergeKit configuration of our approach for a split position $i$. The swapped models exchange the models of the two slices.}
	\label{fig:appendix-mergekit}
\end{figure}

\subsection{Data}
We used the following datasets on the Hugging Face Hub: the \texttt{en-ja\_JP} subset of \texttt{google/wmt24pp} (train split, its only split), the English subset of \texttt{csebuetnlp/xlsum} (test split), \texttt{euclaise/writingprompts} (test split), and \texttt{roskoN/dailydialog} (test split).
We used the texts without preprocessing, except for the following.
For summarization, we kept only the articles with at most 1,024 tokens under the tokenizer of the base model.
The selected articles can therefore differ between LLaMA 2 and Qwen 2.5.
For story generation, the input is the writing prompt of each example without its leading category tag, such as ``\texttt{[ WP ]}''.
For dialogue generation, we prefixed the utterances alternately with ``\texttt{A:~}'' and ``\texttt{B:~}'', joined them with line breaks, and removed the text of the last utterance, so the input ends with the label of the last speaker.
After this processing, we shuffled each split with the \texttt{shuffle} function of Hugging Face Datasets with a seed of 0 and used the first 500 examples.

\subsection{Prompts}\label{sec:appendix-prompts}
All conditions received the plain-text prompts in Table~\ref{tab:appendix-prompts}.
We did not apply chat templates, even to the aligned models.
The tokenizers add their default special tokens to the prompts, such as the BOS token for LLaMA 2.

\begin{table*}[t]
	\centering
	\small
	\begin{tabular}{lp{10cm}}
		\toprule
		Task                & Prompt template                                                                                                \\
		\midrule
		Translation         & \texttt{Translate the following text into English:}\newline\texttt{\{input\}}\newline\texttt{Translation:}     \\
		Summarization       & \texttt{Summarize the following text:}\newline\texttt{\{input\}}\newline\texttt{Summary:}                      \\
		Story generation    & \texttt{Write a story based on the following writing prompt:}\newline\texttt{\{input\}}\newline\texttt{Story:} \\
		Dialogue generation & \texttt{Continue the following dialogue:}\newline\newline\texttt{\{input\}}                                    \\
		\bottomrule
	\end{tabular}
	\caption{Prompt templates. \texttt{\{input\}} is replaced with the input text, and the line breaks are kept as shown.}
	\label{tab:appendix-prompts}
\end{table*}

\subsection{Generation}
We used vLLM~\cite{kwon2023efficient} for generation with the settings in Table~\ref{tab:appendix-generation}.
We generated the $M$ outputs for each input in a single request and used the generated texts without post-processing.
At a temperature of 0.0, we generated one output by greedy decoding and replicated it $M$ times.

\begin{table}[t]
	\centering
	\small
	\begin{tabular}{lL{3.6cm}}
		\toprule
		Setting                & Value                                       \\
		\midrule
		Library                & vLLM 0.28.0                                 \\
		Top-$p$                & 1.0                                         \\
		Top-$k$                & Not applied                                 \\
		Min-$p$                & 0.0                                         \\
		Repetition penalty     & 1.0                                         \\
		Presence penalty       & 0.0                                         \\
		Frequency penalty      & 0.0                                         \\
		Maximum new tokens     & 1,024 (story generation); 512 (other tasks) \\
		Minimum new tokens     & Not applied                                 \\
		Stopping criterion     & EOS token or maximum new tokens             \\
		EOS token IDs          & 2 (LLaMA 2); 151643 and 151645 (Qwen 2.5)   \\
		Maximum context length & 4,096 (LLaMA 2); 32,768 (Qwen 2.5)          \\
		Random seed            & 0                                           \\
		\bottomrule
	\end{tabular}
	\caption{Generation settings. The EOS token IDs are the union of those of the base model and the aligned model, and the maximum context length is the smaller of their values.}
	\label{tab:appendix-generation}
\end{table}

\subsection{Evaluation}
For Self-BLEU, we computed BLEU with $n$-grams up to $n=4$, the mteval-v13a tokenizer, no smoothing, and the shortest reference length for the brevity penalty, as in the default setting of Hugging Face Evaluate.
For the embedding $e(y)$ in Equation~\ref{eq:div_sem}, we used \texttt{Qwen/Qwen3-Embedding-0.6B}~\cite{zhang2025qwen3} with Sentence-Transformers~\cite{reimers2019sentencebert} and no instruction prompt.
For COMET, we used \texttt{Unbabel/wmt22-comet-da}~\cite{rei2022comet22} with the Japanese input as the source.
For BERTScore, we used the F1 score with \texttt{roberta-large}~\cite{liu2019roberta} (layer 17), which is the default model for English, without IDF weighting or baseline rescaling.

\subsection{Software and Hardware}
Table~\ref{tab:appendix-software} lists the versions of the software.
Each run used a single NVIDIA RTX A6000 GPU with 48~GB of memory.

\begin{table}[t]
	\centering
	\small
	\begin{tabular}{lL{3.4cm}}
		\toprule
		Software              & Version                                              \\
		\midrule
		Python                & 3.12                                                 \\
		PyTorch               & 2.13.0 (CUDA 13.0)                                   \\
		Transformers          & 5.16.1 (generation); 4.57.6 (merging and evaluation) \\
		Tokenizers            & 0.23.1                                               \\
		Datasets              & 5.0.1                                                \\
		vLLM                  & 0.28.0                                               \\
		MergeKit              & 0.1.4                                                \\
		Sentence-Transformers & 5.7.0                                                \\
		COMET                 & 2.2.7 (\texttt{unbabel-comet})                       \\
		BERTScore             & 0.3.13 (\texttt{bert-score})                         \\
		Hugging Face Evaluate & 0.4.6                                                \\
		\bottomrule
	\end{tabular}
	\caption{Software versions.}
	\label{tab:appendix-software}
\end{table}

\end{document}